\documentclass[sigconf]{acmart}
\usepackage[font=small,labelfont=bf]{caption}
\usepackage{xspace}
\usepackage{algorithm}
\usepackage{algpseudocode}

\usepackage{enumitem}
\usepackage{graphicx}
\usepackage{subcaption}
\usepackage{enumitem}

\AtBeginDocument{%
  }

\newcommand{\tarag}{TA-RAG\xspace}

\begin{document}

%%
%% The "title" command has an optional parameter,
%% allowing the author to define a "short title" to be used in page headers.
\title{TA-RAG: Tone-Aware Retrieval-Augmented Generation for
Peer-Support Health Communication}

\author{Yong-Bin Kang}
\authornote{Corresponding author.}
\affiliation{%
  \institution{Swinburne University of Technology}
  \city{Melbourne}
  \state{Victoria}
  \country{Australia}
}
\email{ybkang@swin.edu.au}

\author{Anthony McCosker}
\affiliation{%
  \institution{Swinburne University of Technology}
  \city{Melbourne}
  \state{Victoria}
  \country{Australia}
}
\email{amccosker@swin.edu.au}

% \author{Anonymous Author(s)}
% \affiliation{%
%   \institution{Anonymous Institution}
%   \country{}
% }
% \email{}

\newcommand{\yb}[1]{\textcolor{red}{\small\textbf{[YB]} #1 $\triangleleft$}}

% \renewcommand{\shortauthors}{Trovato et al.}

%%
%% The abstract is a short summary of the work to be presented in the
%% article.
\begin{abstract}

Retrieval-augmented generation (RAG) successfully grounds large language model (LLM) outputs in trusted documents, but factual grounding alone is insufficient for sensitive peer-support health communication. In domains such as HIV peer support, responses must also be accessible, stigma-free, empathetic, and tailored to the recipient. This paper presents \tarag, a lightweight, prompt-based tone-aware RAG framework that embeds explicit tone control into a RAG pipeline without requiring model fine-tuning.
We operationalise tone across four core components: stigma-free rewriting, readability adjustment, recipient adaptation, and empathy rephrasing. 
We evaluate \tarag through component-level tests using questions derived from HIV Online Learning Australia (HOLA), UNAIDS terminology guidance, readability metrics, peer-support standards from National Association of People with HIV Australia (NAPWHA), and a public empathy dataset.
Results show that the \tarag's components improve their targeted communication quality while preserving key content. 
These findings emphasise that prompt-based tone control is a potential direction for making RAG outputs suitable for sensitive peer-support health communication.
 
% We position \tarag as an early foundational framework for tone-aware health information systems.

\end{abstract}

\begin{CCSXML}
<ccs2012>
   <concept>
       <concept_id>10010147.10010178.10010179</concept_id>
       <concept_desc>Computing methodologies~Natural language processing</concept_desc>
       <concept_significance>500</concept_significance>
       </concept>
   <concept>
       <concept_id>10010405.10010444.10010449</concept_id>
       <concept_desc>Applied computing~Health informatics</concept_desc>
       <concept_significance>500</concept_significance>
       </concept>
   <concept>
       <concept_id>10002951.10003317.10003338.10003341</concept_id>
       <concept_desc>Information systems~Language models</concept_desc>
       <concept_significance>500</concept_significance>
       </concept>
 </ccs2012>
\end{CCSXML}

\ccsdesc[500]{Computing methodologies~Natural language processing}
\ccsdesc[500]{Applied computing~Health informatics}
\ccsdesc[500]{Information systems~Language models}

\keywords{TA-RAG, Tone-aware Retrieval-Augmented Generation, Retrieval-augmented generation, RAG, peer-support health communication}

%% A "teaser" image appears between the author and affiliation
%% information and the body of the document, and typically spans the
%% page.

\maketitle

\section{Introduction}
Peer-support plays an important role in community health, particularly in contexts where stigma, uncertainty, and health-literacy barriers shape how people seek and interpret information \cite{berg2021, Kang2023Stylometry, who2022,young2024}. In areas such as HIV, mental health, and chronic illness, peer supporters do more than transfer facts—they provide reassurance, practical guidance, and emotionally appropriate communication \cite{hola2023,berg2021,Kang2022Resilience}. This makes peer-support communication a challenging setting for generative AI systems, because a response can be factually correct but still feel clinical, inaccessible, dismissive, or stigmatising.
RAG can provide a promising foundation for health information systems because it grounds generated answers in trusted documents \cite{gao2024rag,Amugongo2025RAGHealthcare}. However, standard RAG pipelines are primarily optimised for relevance and factuality. They do not explicitly control whether the generated response uses person-first language, meets readability needs, adapts to the intended recipient, or acknowledges the user's emotional context \cite{unaids2024,hem2024, Bol:2020,Sharma2023HumanAI}. These limitations are especially important in peer-support settings, where tone can influence trust, comprehension, and willingness to act on information.

This paper asks: \emph{``how can RAG move beyond factual grounding to support tone-aware peer-support health communication?''} To address this question, we propose \tarag, a tone-aware RAG framework that treats tone as an essential design requirement. \tarag defines tone across four dimensions: \emph{stigma-free terminology}, \emph{readability}, \emph{recipient appropriateness}, and \emph{empathy}. The framework adds a tone-adjustment layer after factual draft generation, combined with ambiguity handling. Rather than requiring model fine-tuning or domain-specific model training, \tarag implements these controls through a prompt-based mechanism and guideline-driven rewriting, making the framework lightweight and adaptable for community organisations that may have limited technical resources.

This paper makes three contributions\footnote{Upon acceptance, \tarag's code, prompts, and evaluation datasets will be made available through a GitHub repository.}: First, we propose \tarag, a lightweight architecture for tone-aware RAG in sensitive peer-support health contexts. Second, we operationalise tone through four controllable componnents: stigma-free rewriting, readability adjustment, recipient adaptation, and empathy rephrasing. Third, we evaluate \tarag through component-level testing to validate each module before a full user-deployment study.
The paper is framed as an early foundational framework for \tarag. The goal is to establish whether \tarag's components can produce measurable performance in improving communication quality for peer-supports before broader evaluation with trained peer supporters.

\vspace{-5pt}
\section{Motivation and Tone Definition}

\textbf{{RAG for Health Information Access.}}
RAG systems combine retrieval over  knowledge sources with LLMs, enabling more transparent and grounded responses than purely generative chatbots \cite{gao2024rag,Amugongo2025RAGHealthcare}. In health contexts, RAG is attractive because it can incorporate trusted guidelines, organisational documents, and scientific literature, a capability increasingly explored in medical and health information applications \cite{nori2023capabilitiesgpt4medicalchallenge,Thirunavukarasu2023LLM,Amugongo2025RAGHealthcare}. However, grounding alone does not guarantee that responses are suitable for lay users or sensitive communities. Health information may remain too technical, emotionally flat, or inconsistent with community-preferred language.

\textbf{{Tone in Peer-Support Communication.}}
Health communication research emphasises that effective communication depends on more than correctness \cite{Bol:2020,Mia:2017,Kang2023Stylometry,Lapinski:2025}. Messages need to be readable, respectful, tailored, and emotionally appropriate. In HIV-related communication, for example, terminology can reinforce or reduce stigma, motivating the use of person-first and community-preferred language guidelines \cite{unaids2024,hem2024}. Similarly, the same information may need to be phrased differently for a peer, a general practitioner, or a policymaker, reflecting established work on message tailoring and targeted health communication \cite{Bol:2020,Mia:2017,Lapinski:2025}. \tarag builds on this insight by treating tone not as a superficial style layer, but as a set of operational constraints embedded into the generation pipeline.

We define tone as a composite of four communication properties, which are especially crucial in peer-support health communication:
\begin{enumerate}[leftmargin=*, nosep]
    \item \textbf{Stigma-free terminology}: use of person-first, inclusive, and guideline-aligned language. This dimension is critical because stigmatising or outdated terminology can reduce trust, reinforce social harms, and make health information unsafe \cite{unaids2024,hem2024,who2022}. 
    \item \textbf{Readability}: language that is accessible to non-specialist users. Readability matters because peer-support responses are often used by people with varied health literacy levels, and overly technical language can limit comprehension \cite{Tran2025MedReadCtrl,Bol:2020,Thirunavukarasu2023LLM}.
    \item \textbf{Recipient appropriateness}: adaptation to the communicative expectations of a specified target recipient. Tone is not universal—the same information may need to be conversational and reassuring for a peer, clinically precise for a general practitioner, or concise and impact-oriented for a policymaker \cite{Bol:2020,Mia:2017,Lapinski:2025}.
    \item \textbf{Empathy}: acknowledgement of concern, uncertainty, and emotional context. Peer support often involves emotional reassurance as well as information provision. Prior work on AI-mediated peer support shows that generative models can produce supportive responses. Responses that ignore fear, uncertainty, or distress may appear factual but still fail as supportive communication \cite{Nembhard2023Empathy,Sharma2023HumanAI,young2024,shen2024empathy,Liu2025}.
\end{enumerate}
These properties can reflect specific risks that arise when health information is delivered to people who may be navigating stigma, uncertainty, low health literacy, or emotionally difficult decisions. In this setting, a response must be not only evidence-grounded, but also safe, understandable, socially appropriate, and supportive.
\vspace{-10pt}
\section{TA-RAG Framework}

\tarag extends RAG with a prompt-based tone adjustment layer that ensures responses are stigma-free, readable, empathetic, and appropriate for peer support. 
\tarag takes three main steps (see Fig.~\ref{fig:framework} and also Algorithm~\ref{alg:ta-rag}): 

% (1) clarification and retrieval, (2) draft generation, and (3) prompt-based tone adjustment. 

% The workflow of \textsc{TA-RAG} is described in Algorithm~\ref{alg:ta-rag}:

\begin{figure}[!t]
    \centering
    \includegraphics[width=1\linewidth]{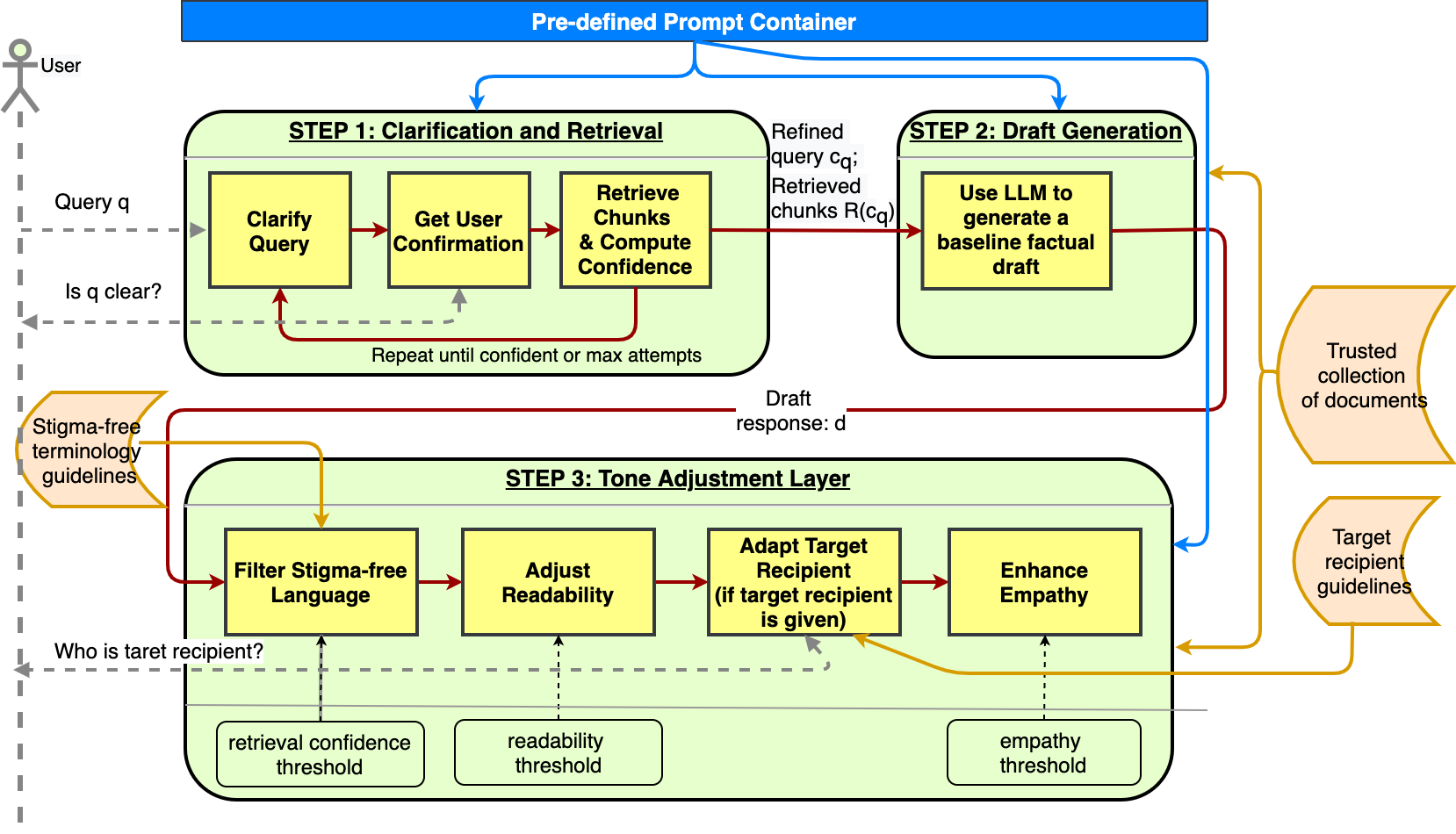}
    \vspace{-8mm} 
    \caption{\tarag architecture. 
    % Its key feature is the tone adjustment layer, which transforms an evidence-grounded RAG draft into a stigma-free, readable, peer-appropriate, and empathetic response.
    }
    \label{fig:framework}
% \vspace{-10pt}
\end{figure}

\setlength{\textfloatsep}{6pt}   % space between float and text
\setlength{\floatsep}{6pt}       % space between floats
\begin{algorithm}[t!]
\small
\caption{TA-RAG Workflow}
\label{alg:ta-rag}
\begin{algorithmic}[1]
\Require Query $q$, corpus $\mathcal{C}$, thresholds $\tau_e$ (empathy), $\tau_r$ (readability), stigma-filtering guidelines $G_s$, recipient  guidelines $G_p$

% \Statex
\Statex \textbf{/* STEP 1: Clarification and Retrieval */}
\State $q \gets \textsc{ClarifyQ}(q, \mathcal{C})$
\State $R(q) \gets \textsc{RetrieveChunk}(k, q, \mathcal{C})$

% \Statex
\Statex \textbf{/* STEP 2: Draft generation */}
\State $d \gets M(q, R(q))$ \Comment{evidence-grounded draft}

% \Statex
\Statex \textbf{/* STEP 3: Tone adjustment layer */}
\State $d' \gets \textsc{Replace}(d, G_s)$ \Comment{stigma-free control}
\If{$R_{read}(d') > \tau_r$} \Comment{readability control}
    \State $d' \gets \textsc{SimplifyUntil}(d', \tau_r)$
\EndIf
\State $d' \gets \textsc{PeerSupportAdapt}(d', G_p)$ \Comment{peer-support alignment}
\If{$E(d') < \tau_e$} \Comment{empathy control}
    \State $d' \gets \textsc{EmpathyRephrase}(d')$
\EndIf

\State \Return $d'$
\end{algorithmic}
\end{algorithm}

\textbf{STEP 1: Clarification and Retrieval:}
\tarag begins to clarify the user query $q$ (or prompt). A prompt-based ambiguity detector checks whether $q$ is clear enough to be answered based on  the corpus $\mathcal{C}$ (\textbf{L1}). If $q$ is ambiguous, a more clarified question is generated and presented to the user for confirmation or refinement. For example, a query \textit{``How can I improve my quality of life?''} is broad,  as quality of life for people living with HIV may refer to social connection,  distress, or day-to-day support. In the NAPWHA peer-support context, \tarag may generate a clarifed  query: \textit{``Which part of living well with HIV would you like support with: connection, emotional wellbeing, healthcare, stigma, or distress?''}
Once $q$ is confirmed by the user to be clear, \tarag retrieves the top-$k$ ($k$: user-specified) relevant chunks from $\mathcal{C}$ using $\textsc{RetrieveChunk}(\cdot)$ (\textbf{L2}):
$
\operatorname{top\text{-}k}_{p \in \mathcal{C}} \cos(\operatorname{em}(q), \operatorname{em}(p)),
$
where $\operatorname{em}(\cdot)$ is the embedding function, and $p$ is a document chunk. The retrieved chunks provide the factual grounding for the response.

\textbf{STEP 2: Draft Generation:}
The LLM $M$ generates an evidence-grounded factual draft $d$ from  $\mathcal{C}$, conditioned on $q$ and the retrieved chunks $R(q)$  (\textbf{L3}).
The draft $d$ is intended to maximise factual relevance and coverage by drawing on $\mathcal{C}$. Nevertheless, it may still be too complex, impersonal, or misaligned with the supportive, accessible, and communication norms expected by the user. 
% Thus, in \tarag, $d$ is treated as an intermediate response rather than the final user-facing response.

\textbf{STEP 3: Tone Adjustment Layer:}
This layer refines the draft $d$ through four prompt-based components using the LLM $M$  (\textbf{L4--11}):

\textbf{Stigma-free rewriting ($\mathbf{C}_{\mathbf{Stig}}$):} 
$\mathbf{C}_{\mathrm{Stig}}$ takes $d$ as input and transforms  stigmatised expressions into preferred alternatives  using the rules derived from  stigma-filtering guidelines $G_s$ (\textbf{L4}). For example, in HIV peer-support contexts, $G_s$ can be operationalised as a domain-specific stigma-filtering rule set derived from \textit{UNAIDS Terminology Guidelines}~\cite{unaids2024} (e.g., \textit{`HIV patients}' $\rightarrow$ \textit{`people living with HIV}'). By replacing labeling language while preserving semantic meaning, $\mathbf{C}_{\mathrm{Stig}}$ produces a refined draft $d'$, which subsequently serves as the input for the downstream tone-control steps. 

\textbf{Readability control ($\mathbf{C}_{\mathbf{Read}}$):} $\mathbf{C}_{\mathrm{Read}}$ evaluates the readability of $d'$ using $R_{read}(\cdot)$ and compares it against the  threshold $\tau_r$ (\textbf{L5--7}). If $R_{read}(d') > \tau_r$, the response is more complex than the desired readability level, $\mathbf{C}_{\mathbf{Read}}$ invokes $\textsc{SimplifyUntil}(d', \tau_r)$ to iteratively refine and rewrite the response until it satisfies the condition, or the user-specified maximum $\#$ of simplification attempts is reached. 

\textbf{Recipient adjustment ($\mathbf{C}_{\mathbf{Reci}}$):} $\mathbf{C}_{\mathbf{Reci}}$ adapts $d'$ to the intended recipient using recipient guidelines $G_p$ (\textbf{L8}). The user may specify a recipient category, such as peer-supporter or policymaker, where $G_p$ defines the expected formality and communicative style. A policymaker-facing response may prioritise policy relevance, whereas a peer-support response may emphasise reassurance, shared understanding, and non-judgemental language. $\mathbf{C}_{\mathbf{Reci}}$ updates $d'$ to align it with the communicative norms of the intended recipient.

\textbf{Empathy rephrasing ($\mathbf{C}_{\mathbf{Emph}}$):} $\mathbf{C}_{\mathbf{Emph}}$ evaluates $d'$ using an empathy score $E(\cdot)$ and compares it with the empathy threshold $\tau_e$ (\textbf{L9--11}). If $E(d') < \tau_e$, \tarag refines $d'$ with empathetic framing that acknowledges concern, uncertainty, or emotional context.  The rewritten response may acknowledge the user's emotions and offer reassurance, for example, \textit{``I understand this may feel overwhelming, and you are not alone in this.''} This approach supports flexible, domain-adapted empathy control.
% These integrated components define \tarag as a lightweight to standard RAG pipelines, where output tone is refined across safety, accessibility, audience alignment, and empathy. 

The integration order in \textbf{STEP 3} is deliberate: $\mathbf{C}_{\mathbf{Stig}}$ first removes stigmatising terminology; $\mathbf{C}_{\mathbf{Read}}$ then improves readability; $\mathbf{C}_{\mathbf{Reci}}$ aligns the response with the intended audience; and $\mathbf{C}_{\mathbf{Emph}}$ is applied last so that empathetic framing is preserved in the final response. Alternative orderings could  be evaluated in future work.

\vspace{-15pt}
\section{Evaluation Approach}

We evaluate \tarag through component-level testing. Because \tarag is designed as a modular architecture, assessing each component separately provides the evidence of its purposed tone-aware  generation, while preserving the factual RAG response. All four components are implemented using prompt-based inference\footnote{All prompts used in the evaluation will be released in our GitHub repository upon acceptance.}. Unless otherwise noted, {`gpt-4o-mini'} is used as the default LLM for response and sample question generation. The RAG corpus $\mathcal{C}$ is built from the \textit{HOLA quality of life report}~\cite{hola2023}, which guides models for improving quality of life of people with HIV in Australia. 

\textbf{Evaluation of $\mathbf{C}_{\mathbf{Stig}}$:}
The stigma-filtering guidelines $G_s$ is instantiated from \textit{UNAIDS Terminology Guidelines}~\cite{unaids2024}.
We evaluate $\mathbf{C}_{\mathbf{Stig}}$ using a synthetic dataset $D_{\mathrm{trig}}$ derived from $\mathcal{C}$ and $G_s$.  For each of the 105 rules in $G_s$, we use the LLM to generate 5 sentences containing the discouraged expression in the rule, thus $|D_{\mathrm{trig}}|=525$.
We calculate three metrics: 1) \textit{ReplaceRate}, the ratio of  examples in $D_{\mathrm{trig}}$, where each  example $d$ with a discouraged expression is replaced by a sentence $d'$ with the recommended expression defined in $G_s$; 2) \textit{SemSim},  the cosine similarity between the vector representations of $d$ and $d'$ using {`text-embedding-3-small'}; 3) $\textit{ContPreserve}$ uses \textit{BERTScore Recall} \cite{zhang2020bertscore} to assess if key concepts in $d$ are preserved in $d'$. 
While \textit{SemSim} captures overall semantic similarity, \textit{ContPreserve} provides a finer-grained check of content preversation. Both \textit{SemSim}($\cdot$) and \textit{ContPreserve}($\cdot$) are also used in the remaining component evaluations to measure semantic preservation.

\textbf{Evaluation of $\mathbf{C}_{\mathbf{Read}}$:}
We evaluate $\mathbf{C}_{\mathbf{Read}}$ to test whether it can improve readability while preserving semantic meaning. 
Inspired by the work on LLM-based automated question generation~\cite{Kang:2025}, we generate 100 questions from $\mathcal{C}$ and use the LLM to produce corresponding responses $S$.
We then apply $\mathbf{C}_{\mathbf{Read}}$ to each response $d \in S$ to produce a readability-adapted response $d'$. We use a \textit{Grade 8} level as the target readability level for our testing.
We evaluate three metrics: 1) \textit{Readability} measures whether $d'$  falls within the Grade 8 level, using two complementary metrics:
\textit{Flesch--Kincaid (FK)}~\cite{Challener2025FKGL}, which captures surface-level readability through surface features (e.g., syllable count, word/sentence length), and \textit{zero-shot LLM-based Automatic Readability Assessment (ARA)}~\cite{grossman:2026}, which estimates contextual readability from an LLM's probability distribution over difficulty scores. The ARA uses a 1--5 scale: \textit{1: elementary}, \textit{2: middle school}, \textit{3: high school}, \textit{4: college}, and \textit{5: academic}. The Grade 8 level roughly corresponds to $\mathrm{ARA}\approx 2$, or the lower end of 3; (2) \textit{SemSim} and (3) \textit{ContPreserve}, both  computed between $d$ and $d'$. 
% \textit{1: elementary/very easy}, \textit{2: middle school/easy}, 3: \textit{high school/moderate}, \textit{4: college/difficult}, and \textit{5: academic/extremely dense}. 

\textbf{Evaluation of $\mathbf{C}_{\mathbf{Reci}}$:}
We evaluate $\mathbf{C}_{\mathbf{Reci}}$ using the same 100 questions and their non-readability-adapted responses $S$ used in the $\mathbf{C}_{\mathbf{Read}}$ evaluation. For each response $d \in S$, $\mathbf{C}_{\mathrm{Reci}}$ produces a peer-support-adapted response $d'$, based on $G_p$ denoting peer-support communication guidelines: \textit{NAPWHA Australian Peer Support Standards}~\cite{napwha2020peerstandards}.
We measure three metrics: 1) \textit{PeerAlign} measures whether $d'$ is appropriate for peer-support communication, using three LLM-as-rater models (i.e., `mistral-small-2603', `deepseek-v4-flash', `qwen3.6-plus-2026-04-02'),  providing each model with $G_p$ as evaluation criteria. The models assign a 1--5 score based on whether the response is suitable for peer-to-peer communication; 2) \textit{SemSim} and (3) \textit{ContPreserve}, both computed between $d$ and $d'$.

\textbf{Evaluation of $\mathbf{C}_{\mathbf{Emph}}$:}
To evaluate $\mathbf{C}_{\mathbf{Emph}}$, we use the \textit{Empathy in Text-based Mental Health Support} dataset\footnote{\url{https://github.com/behavioral-data/Empathy-Mental-Health}} and select 100 strongly empathetic Reddit responses (empathy level=2). Given each of these responses $d$, we generate a non-empathetic version $d'$ using the LLM to simulate a baseline response. We then apply $\mathbf{C}_{\mathbf{Emph}}$ to produce an empathetic response $d''$ from $d'$.
We evaluate three metrics: (1) \textit{EmScore}, which measures the empathetic quality of $d''$ using the same three LLM-as-rater models, as the evaluation of $\mathbf{C}_{\mathbf{Reci}}$, comparing with $d'$. Each LLM assigns a 1--5 score based on whether the response reflects empathetic framing; (2) \textit{SemSim} and (3) \textit{ContPreserve} are computed between $d$ and $d''$. 

\textbf{Parameter Setting:} $\tau_r$ and $\tau_e$ can be initialised using heuristics and refined via human feedback (e.g., Likert-style user perception rubrics). Optimising such rubrics is outside the scope of this work.
\vspace{-5pt}
\section{Evaluation Results}

\begin{figure}[!h]
    \centering
    \subcaptionbox{ReplaceRate Score\label{fig:b}}{%
        \includegraphics[width=0.28\columnwidth, height=1in]{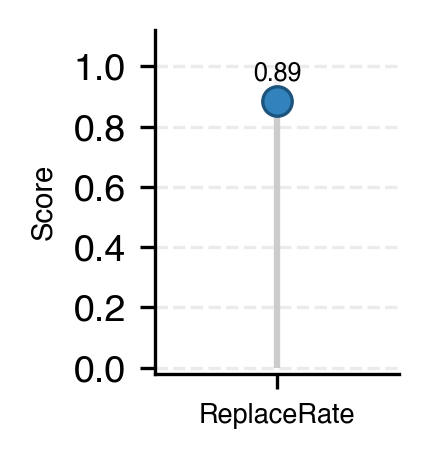}%
    }
    % \hfill
    \hspace{20pt}
    \subcaptionbox{Semantic Preservation Dist.\label{fig:a}}{%
        \includegraphics[width=0.4\columnwidth, height=1in]{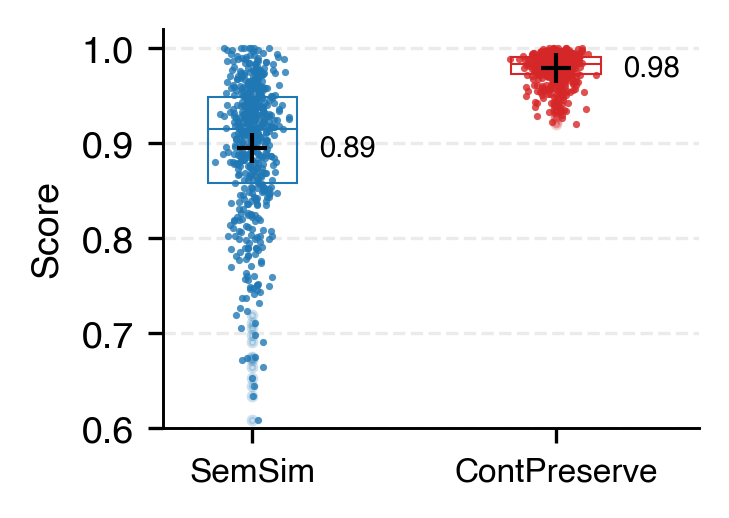}%
    }%
    \vspace{-10pt}
    \caption{Stigma-filtering component evaluation}
    \label{fig:stigma}
 \vspace{-5pt}
\end{figure}

\textbf{Evaluation of $\mathbf{C}_{\mathbf{Stig}}$:}
Fig.~\ref{fig:stigma} (a) shows that $\mathbf{C}_{\mathrm{Stig}}$ achieves  $\textit{ReplaceRate}=0.89$. The  0.11 gap is mainly due to the exact-match strictness of \textit{ReplaceRate}: 1) $D_{\mathrm{trig}}$ often produces a contextually relevant paraphrase instead of the rule-based replacement in $G_s$: e.g., given a rule: $\textit{addict} (r_1) \rightarrow \textit{person who uses drugs} (r_2)$, $\mathbf{C}_{\mathrm{Stig}}$ transformed  \textit{`an addict for ten years'} into \textit{`struggling with substance use for ten years'};  here $r_1$ is removed, but the output does not contain exact $r_2$; (2)
Some preferred expressions contain the discouraged term as a substring: e.g., given,  \textit{sex worker}($r_1$) $\rightarrow$ \textit{commercial sex worker}($r_2$), the output is counted as an error, as $r_1$ appears inside $r_2$.
Thus, the errors partly stem from the exact matching, not failures in stigma reduction. Fig.~\ref{fig:stigma} (b) shows \textit{SemSim}=0.89; \textit{ContPreserve}=0.98, indicating $\mathbf{C}_{\mathrm{Stig}}$ strongly retains both the overall semantic meaning and finer-grained key concepts from the original sentence in the rewritten output. Overall, $\mathbf{C}_{\mathbf{Stig}}$  removes or softens discouraged terms while largely preserving the semantic content of the input.

\begin{figure}[!h]
    \centering
    \subcaptionbox{Readability Comparison\label{fig:b}}{%
        \includegraphics[width=0.48\columnwidth, height=1.2in]{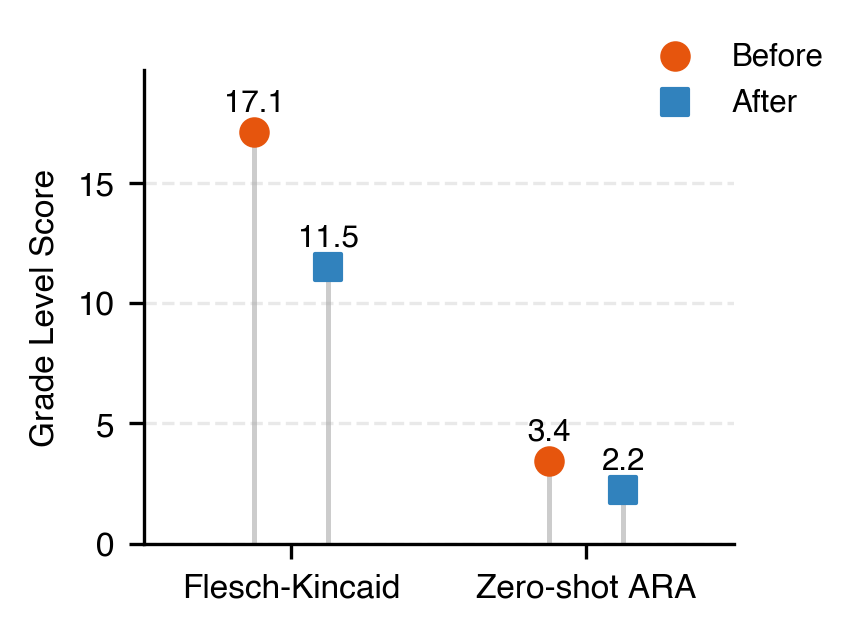}%
    }
    \hspace{10pt}
    \subcaptionbox{Semantic Preservation Dist.\label{fig:a}}{%
        \includegraphics[width=0.42\columnwidth]{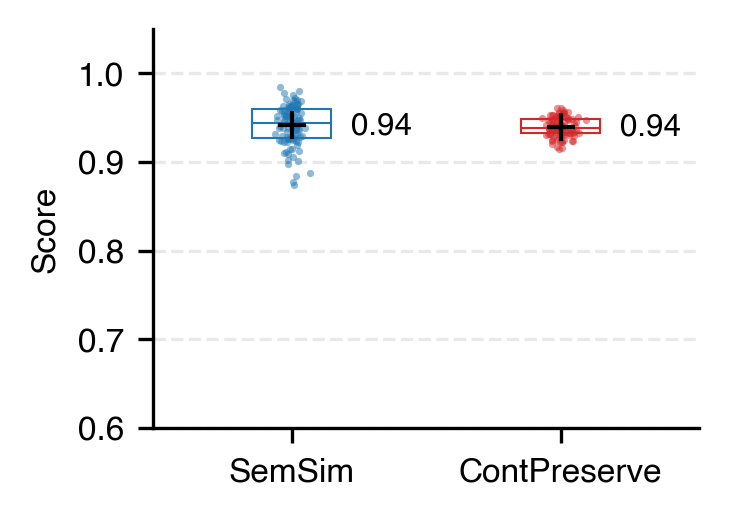}%
    }%
    \vspace{-10pt}
    \caption{Readability component evaluation}
    \label{fig:read}
 \vspace{-5pt}
\end{figure}

\textbf{Evaluation of $\mathbf{C}_{\mathrm{Read}}$:}
% a zero-shot LLM-based Automatic Readability Assessment (ARA) metric~\cite{grossman:2026}. ARA uses a 1--5 difficulty scale, where \textit{1: elementary/very easy}, \textit{2: middle school/easy}, 3: \textit{high school/moderate}, \textit{4: college/difficult}, and \textit{5: academic/extremely dense}. Since our target is Grade 8 readability, ARA scores around 2, or the lower end of 3, indicate closer alignment with the intended readability level.
Fig.~\ref{fig:read}(a) shows that $\mathbf{C}_{\mathrm{Read}}$ reduces reading complexity under both metrics. 
% But these metrics capture different aspects of simplification. 
FK decreases from 17.1 to 11.5, but remains above the Grade 8 as it is driven by surface features, especially syllables per word. 
Health-related terms often contain multiple syllables (e.g, HIV, medication, {antiretroviral}), which can keep FK scores high even when a text is conceptually easier. 
In contrast, the ARA score decreases from 3.4 to 2.2, shifting the response from an upper secondary-level towards a middle-school level. 
This contrast is illustrated by the example pair: the initial response \textit{``Antiretroviral adherence reduces morbidity in HIV patients}'' is rated as difficult by both metrics (FK=17, ARA=4.6). The simplified output \textit{``Taking HIV medication reduces illness in patients}'' still receives FK=12 (as it retains multi-syllable health-related terms: e.g., \textit{HIV}, \textit{medication}), but its ARA score decreases to 2.22. This indicates that, in health-related contexts, formula-based readability scores should be interpreted cautiously and complemented with contextual readability estimates.
Fig.~\ref{fig:read}(b) shows strong meaning preservation, with both \textit{SemSim} and \textit{ContPreserve}=0.94. 
Overall, $\mathbf{C}_{\mathrm{Read}}$ improves readability while preserving content.

\begin{figure}[!h]
    \centering
    \subcaptionbox{Recipient Score Comparison\label{fig:b}}{%
        \includegraphics[width=0.48\columnwidth, height=1.2in]{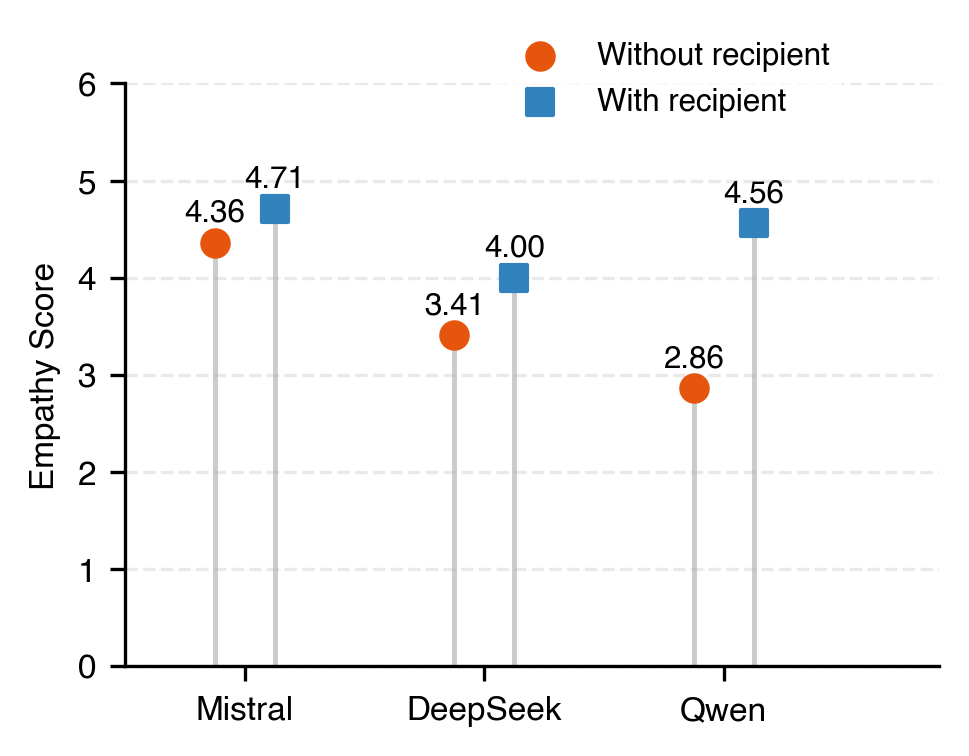}%
    }
    \hfill
    \subcaptionbox{Semantic Preservation Dist.\label{fig:a}}{%
        \includegraphics[width=0.42\columnwidth]{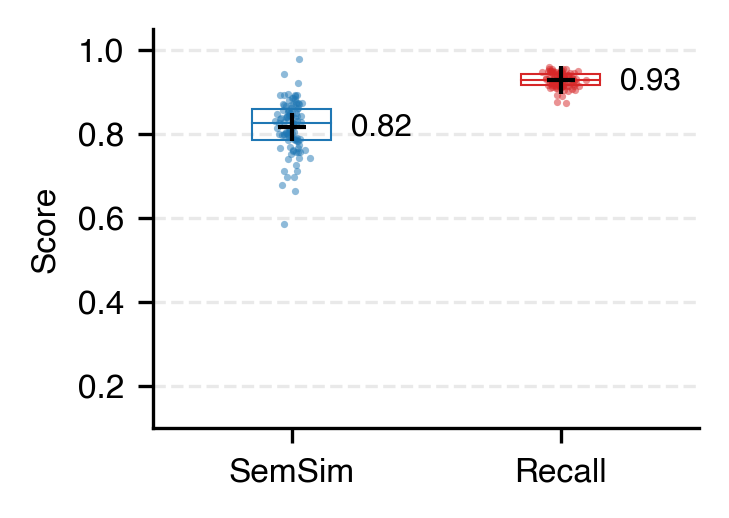}%
    }%
        \vspace{-10pt}
    \caption{Recipient adjustment component evaluation}
    \label{fig:recipient}
        \vspace{-5pt}
\end{figure}

\textbf{Evaluation of $\mathbf{C}_{\mathbf{Reci}}$:} Fig.~\ref{fig:recipient}(a) shows that recipient guidance improves peer-support alignment across all three LLM-as-rater models. Mistral increases from 4.36 to 4.71, DeepSeek from 3.41 to 4.00, and Qwen from 2.86 to 4.56. Averaged across the three raters, the alignment score increases from 3.54 to 4.42, with an average gain of 0.88. 
% The largest gain is observed for Qwen. 
The results indicate that explicit recipient guidance is helpful when the baseline response is less naturally aligned with peer-support norms. Differences in LLM-as-rater scores likely reflect model-specific calibration; however, the consistent improvement across all raters demonstrates that $\mathbf{C}_{\mathrm{Reci}}$ is robust to evaluator variation.
Fig.~\ref{fig:recipient}(b) shows that recipient adaptation preserves content reasonably well, with \textit{SemSim}=0.82 and \textit{Recall}=0.93. The \textit{SemSim} score reflects a high stylistic shift introduced by $\mathbf{C}_{\mathrm{Reci}}$, which adapts the response toward peer-support communication. The higher recall score nevertheless indicates that key concepts from the original response are retained.

\begin{figure}[!h]
    \centering
    \subcaptionbox{\textit{EmScore} Comparison\label{fig:b}}{%
        \includegraphics[width=0.48\columnwidth, height=1.3in]{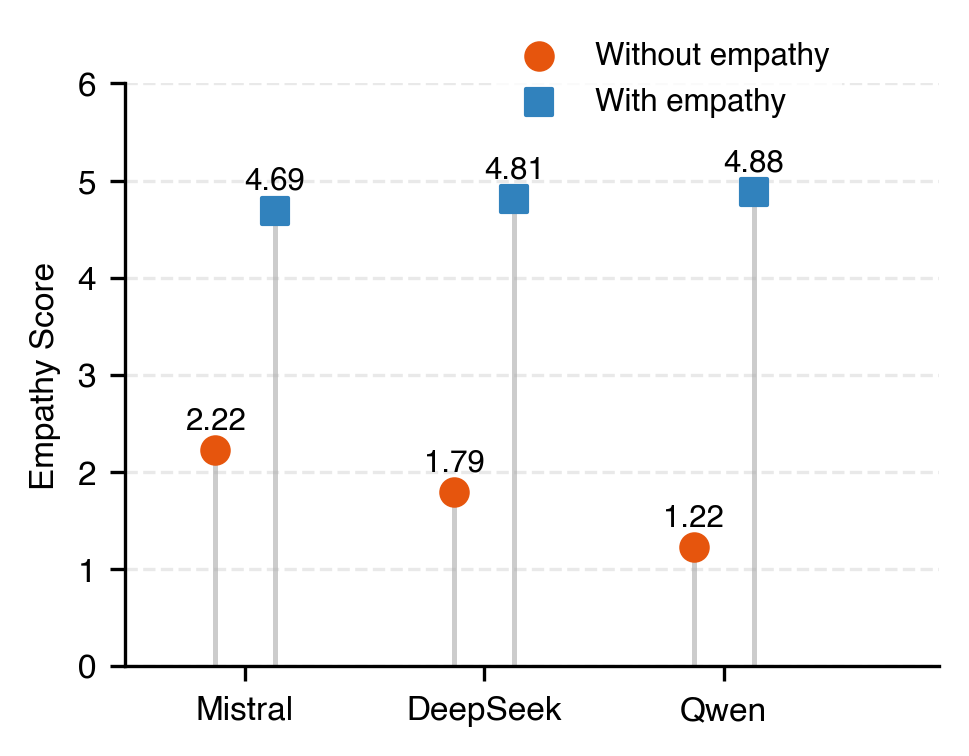}%
    }
    % \hfill
    \subcaptionbox{Semantic Preservation Dist.\label{fig:a}}{%
        \includegraphics[width=0.42\columnwidth]{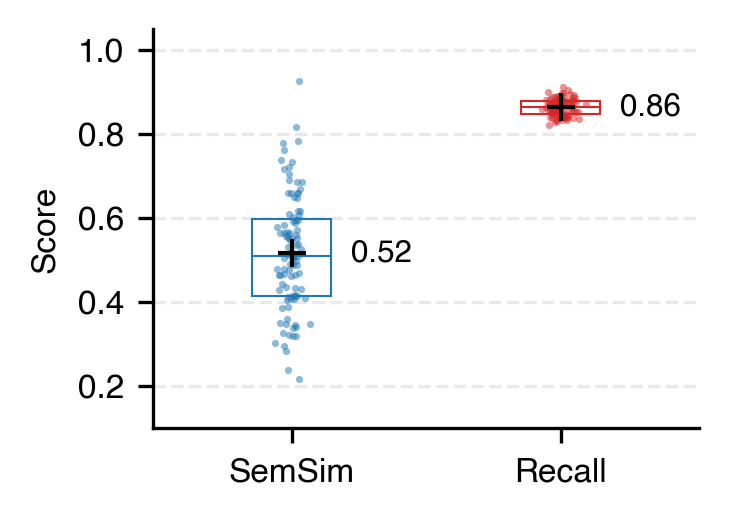}%
    }%
    \vspace{-10pt}
    \caption{Empathy rephrasing component evaluation}
    \label{fig:empathy}
        \vspace{-10pt}
\end{figure}

\textbf{Evaluation of $\mathbf{C}_{\mathbf{Emph}}$:}
Fig.~\ref{fig:empathy}(a) shows that $\mathbf{C}_{\mathrm{Emph}}$ substantially improves empathy scores across the rating models: Mistral  from 2.22 to 4.69, DeepSeek from 1.79 to 4.81, and Qwen from 1.22 to 4.88. Averaged, the score increases from 1.74 to 4.79 (a 3.05 gain), indicating that empathy rephrasing consistently shifts responses toward empathetic communication.
Fig.~\ref{fig:empathy}(b) shows \textit{SemSim}=0.52, with a higher \textit{Recall}=0.86. This pattern is explained by the scale of rewriting: $\mathbf{C}_{\mathrm{Emph}}$ produces longer outputs, with a mean relative length (MRL) increase of $1.69 \pm 2.23$, indicating that it adds long contextual and empathetic elaboration rather than only making local edits. This expansion is negatively correlated with \textit{SemSim} ($r=-0.59$), suggesting that larger stylistic and explanatory additions reduce \textit{SemSim}. The \textit{Recall} score of 0.86 still indicates a high degree of content preservation, although it is lower than in other components because the rewritten output may no longer align token-by-token or phrase-by-phrase with the original.
Compared with stigma filtering, readability adjustment, and recipient adaptation, which show smaller MRL changes ($0.08\pm0.14$, $-0.14\pm0.09$, and $0.17\pm0.49$) and higher \textit{SemSim} scores (0.89, 0.94, and 0.82), $\mathbf{C}_{\mathrm{Emph}}$ produces the strongest stylistic expansion. Overall, the results show that $\mathbf{C}_{\mathrm{Emph}}$ improves empathetic framing while preserving core content.

\vspace{-5pt}
\section{Conclusion} \label{sec:conclusion}
This paper proposes \tarag, a lightweight, prompt-based tone-aware RAG framework for adding explicit tone control to RAG in sensitive peer-support health contexts. \tarag decomposes tone into four prompt-based components: stigma-free rewriting, readability adjustment, recipient adaptation, and empathy rephrasing. The component-level evaluation shows that these components improve their targeted communication properties while preserving key content across all components, with content-preservation scores remaining high, ranging from 0.86 to 0.98.
The results also reveal a key trade-off: local edits, such as stigma filtering and readability adjustment, preserve semantic similarity more strongly, while more generative edits, such as recipient adaptation and empathy rephrasing, introduce larger stylistic shifts but retain key content. Future work will evaluate \tarag end-to-end with trained peer supporters and target users in realistic peer-support scenarios.

\begin{acks}
We thank Bay Nandavong for contributing to implementing the evaluation code and generating the experimental outputs used in this study.
\end{acks}

\section*{GenAI Usage Disclosure}
In preparing this paper, GPT-4o was used for identifying and correcting grammatical errors, typos, and other writing mistakes. In accordance with academic integrity, all research content, methods, and data analysis were developed, conducted, and validated by the authors.
%%
%% The next two lines define the bibliography style to be used, and
%% the bibliography file.
\bibliographystyle{ACM-Reference-Format}
\bibliography{tarag}

\end{document}